\documentclass{article}

\usepackage[preprint]{neurips_2026}
\usepackage[utf8]{inputenc}
\usepackage[T1]{fontenc}
\usepackage{hyperref}
\usepackage{url}
\usepackage{booktabs}
\usepackage{amsmath}
\usepackage{amsfonts}
\usepackage{amssymb}
\usepackage{microtype}
\usepackage{xcolor}
\usepackage{graphicx}
\usepackage{tabularx}
\usepackage{array}
\usepackage{enumitem}
\usepackage{caption}
\usepackage{multirow}

\graphicspath{{figures/}}
\setlist{leftmargin=*,nosep}
\newcolumntype{Y}{>{\raggedright\arraybackslash}X}

\newcommand{\C}{\mathcal{C}}

\hypersetup{
  colorlinks=true,
  linkcolor=blue!55!black,
  citecolor=blue!55!black,
  urlcolor=blue!55!black
}

\title{Same Targets, Different Computation:\\
How Post-Training Divides Work Across Model Layers}

\author{Yifan Zhou\\
University of California, Los Angeles\\
\texttt{yifanz1207@gmail.com}\\
\href{https://github.com/yifan1207/first-divergence-crosspatching}{github.com/yifan1207/first-divergence-crosspatching}}

\begin{document}
\maketitle

\begin{abstract}
A late-layer change learned during post-training may work on the base model's earlier state, or it may depend on earlier computation learned alongside it. We distinguish these cases with a four-cell diagnostic that crosses base or descendant upstream states with base or descendant late stacks. One might expect that the larger a learned late change, the more it needs the earlier computation that came with it. It does not: OpenMath2's late stack moves the math-target margin by $+3.43$ logits even after the \emph{base} model's earlier state, giving a near-zero interaction, while instruction-following descendants of the same Llama-3.1-8B base depend heavily on their own. Across one fixed support, seven released descendants span $-0.54$ to $+2.20$ logits, and controlled code and biomedical continuation-training runs sit near zero. Released checkpoints differ in many ways, so we then isolate one training property. Two LoRA fine-tunes learn identical target responses, requested either by familiar natural-language instructions or by newly learned nonce codes. Changing only this cue-to-response relation increases upstream dependence by $+5.56$ logits on Qwen3-4B and $+4.18$ on Llama-3.1-8B, positive in all six model-by-seed runs. The interaction is positive in all five released base/instruction pairs we test, and late-stack replacement changes the full-vocabulary argmax in about half of events. Post-training can therefore organize the same target behavior differently across depth, even with the base model and the targets held fixed.
\end{abstract}

\section{Introduction}
\label{sec:intro}

Fine-tuning changes what a language model does. It can also change \emph{how} the model divides a computation across depth. A late-layer change may be useful on its own: replacing the base model's late stack could produce the learned output even when it receives the base model's earlier-layer state. Or the same late change may work mainly inside its descendant checkpoint because earlier layers prepare the state it expects. Standard output evaluations do not distinguish these cases, and localization inside one checkpoint does not distinguish them either.

This distinction matters for model diffing. Recent work has localized post-training changes to directions, features, and layers, or has asked whether fine-tuning strengthens a circuit already present in the base model \citep{prakash2024finetuning,du2025posttraining,bigoulaeva2026patches}. Such analyses identify where a difference can be read out or controlled. They do not by themselves tell us whether the localized change survives when the computation before it comes from the paired base model. That missing comparison is especially important when a late component is interpreted as a self-contained mechanism.

\begin{figure}[t]
  \centering
  \includegraphics[width=\linewidth]{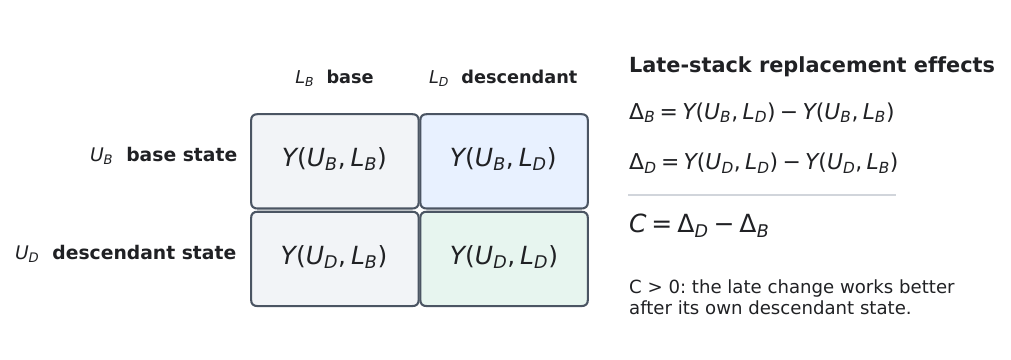}
  \caption{\textbf{Cross-patching measures whether a descendant late-stack change depends on descendant upstream state.} The same late-stack replacement is evaluated after base state ($\Delta_B$) and descendant state ($\Delta_D$). Their difference $\C$ is positive when the late change works better after its own descendant's earlier computation.}
  \label{fig:diagnostic}
\end{figure}

We make that comparison directly. Given a base checkpoint $B$ and a descendant $D$ with the same architecture, we cross the state entering a fixed late boundary with the late stack from either checkpoint. We measure the descendant-target margin in all four combinations. The difference between the late-stack replacement effect after descendant state and the same replacement effect after base state is an \emph{upstream-dependence interaction}. A positive interaction means that the learned late change is more effective after its own descendant's earlier computation. Figure~\ref{fig:diagnostic} shows the test.

A natural expectation is that the larger a learned late change, the more it should need the earlier computation that came with it. Our first result runs against that expectation. OpenMath2's late stack moves the math-target margin by more than three logits even when the state feeding it comes from the base model, which never received the math fine-tune, so a large learned effect can be almost indifferent to what produced its input. Several instruction-following descendants of the same base do the opposite, depending heavily on their own earlier computation, while code and biomedical continuation-pretraining (CPT) controls sit near zero on the common support despite measurably learning their domains. On one fixed prompt support, boundary, and readout, seven released descendants of Llama-3.1-8B span interactions from $-0.54$ to $+2.20$ logits. Their common origin rules out architecture and neuron indexing as explanations for this spread, but not differences in data or training. Something about the training itself sets the profile, which motivates a controlled test of whether the learned cue-to-response relation can shift it.

Released models still confound data, objective, update size, and training history. We therefore run a controlled experiment with the same base models, source prompts, target responses, response multiset, optimizer, LoRA configuration, and update count. The only change is how one of four output modes is requested. In the \emph{familiar} arm, an ordinary instruction requests plain text, JSON, XML, or a reversed answer. In the \emph{novel} arm, a fixed nonce code requests the identical response. Learning the new cue relation shifts the interaction toward greater own-state dependence by $+5.56$ logits on Qwen3-4B and $+4.18$ on Llama-3.1-8B. The shift is positive for all three training seeds on both bases. Thus identical target responses can be learned with different dependencies between earlier and later computation.

We then test breadth and consequences. Across five released base/instruction pairs from 4B to 32B, the interaction is positive in every family. The late replacement changes the full-vocabulary top-1 token in roughly half of selected events. On GSM8K, forcing the descendant-preferred token and continuing with the native descendant changes suffix-only exact-answer accuracy by $+3.2$ and $+2.5$ percentage points in two descendants. These results show that the local quantity reaches actual token decisions and that those decisions can matter downstream.

The practical consequence is one additional comparison. A study that localizes a post-training effect to late layers should report that late replacement under \emph{both} the descendant's and the paired base model's upstream state. A large native effect $\Delta_D$ shows only that the late change matters inside its own checkpoint. The pair $(\Delta_B,\C)$ separates a late change that already works on base state from one whose native effect depends on earlier descendant computation. Once cross-model patching is available, this comparison requires no new model training and prevents late \emph{expression} from being read as late \emph{self-containment}.

Our contributions are:
\begin{enumerate}
  \item a four-cell cross-patching diagnostic that measures whether a learned late-stack effect works on base upstream state or depends on its descendant's earlier computation;
  \item a coherent same-base atlas and a controlled same-target experiment showing that post-training can produce different upstream--late dependency profiles; and
  \item breadth across five model families, decision-level effects, and focused controls for alignment, boundary choice, token selection, and history construction.
\end{enumerate}

\section{Measuring Dependence Between Earlier and Later Computation}
\label{sec:method}

\subsection{The four-cell diagnostic}

The question in one sentence is: \emph{does the descendant's late stack retain its learned effect when it receives the base model's earlier state?} We answer it by measuring the same late-stack swap after each model's earlier computation and taking the difference.

Let $B$ be a base checkpoint and $D$ a same-architecture descendant. At a fixed boundary, $U_B$ and $U_D$ are the residual states produced by the layers before the boundary, and $L_B$ and $L_D$ are the layers from the boundary through the final transformer block. For a scored target $t^+$ and comparator $t^-$, define
\begin{equation}
Y(U,L)=\operatorname{logit}_{U,L}(t^+) - \operatorname{logit}_{U,L}(t^-).
\end{equation}
We run all four combinations $(U_B,L_B)$, $(U_B,L_D)$, $(U_D,L_B)$, and $(U_D,L_D)$. The descendant late-stack replacement effects are
\begin{align}
\Delta_B &= Y(U_B,L_D)-Y(U_B,L_B),\\
\Delta_D &= Y(U_D,L_D)-Y(U_D,L_B).
\end{align}
Their difference,
\begin{equation}
\C=\Delta_D-\Delta_B,
\label{eq:interaction}
\end{equation}
is the upstream-dependence interaction. If $\C\approx 0$, the measured late change is similarly effective after base and descendant upstream states. If $\C>0$, it is more effective after the descendant's state. The quantity is a local dependency measure, not the total effect of fine-tuning and not a claim that one training stage caused a particular layer to change.

For released checkpoint pairs, we start from a raw prompt and compare greedy top-1 predictions step by step. The prefix is identical until the first disagreement by construction. That position supplies $t^+$ (the descendant token) and $t^-$ (the base token). We use first divergence because it is the earliest naturally occurring position at which the two checkpoints define a directional next-token contrast. We do not treat it as an average token.

In the primary readout, all four cells use the descendant's final normalization, output head, and vocabulary mask; repeating the analysis with the base model's readout is a secondary check. Each hybrid is recomputed as a full sequence forward pass. The prefix cache is used only to locate the disagreement; it is not spliced across models. At evaluation, the hidden state entering the late boundary is replaced for all sequence positions and the downstream stack recomputes its own keys and values. Patching a model's own state back into itself reproduces its native logits exactly, which confirms the machinery is faithful. Appendix~\ref{app:implementation} gives the implementation details.

\subsection{Released-checkpoint supports}

The same-base atlas uses Llama-3.1-8B as $B$ and seven released descendants as $D$: Meta Instruct, Hermes-3, Tulu SFT, Tulu DPO, Tulu Final, DeepSeek-R1-Distill-Llama, and OpenMath2 \citep{teknium2024hermes3,deepseek2025r1,lambert2024tulu3,toshniwal2024openmath}. Every point uses the same 1,400-prompt manifest, boundary at layer 19 of 32, first-divergence rule, and common-descendant readout. The Tulu checkpoints are cumulative stages from one lineage and are shown as a connected series, not as independent models or a causal stage decomposition. Two controlled CPT adapters, trained on code or biomedical text from the same base, are evaluated beside the released atlas.

For breadth, the Core-5 analysis uses released base/instruction pairs from Llama-3.1-8B, Qwen3-4B, Mistral-7B-v0.3, OLMo-2-7B, and Qwen2.5-32B. Late boundaries were fixed at roughly 60\% depth before we ran the combined analysis, not tuned afterwards. The common holdout contains 600 prompts spanning conversation, explicit formatting, and safety. Appendix~\ref{app:models} lists model revisions, boundaries, prompt composition, and valid-event counts.

\subsection{Controlled same-target experiment}
\label{sec:controlled-method}

Released checkpoints cannot attribute their different interactions to a single training property. We therefore construct two response-only LoRA fine-tunes \citep{hu2022lora} on Qwen3-4B-Base and Llama-3.1-8B. Every source problem appears in four target modes: plain answer, one-field JSON, XML, and reversed answer. The \emph{familiar} arm specifies the mode in ordinary language. The \emph{novel} arm replaces that instruction with a stable nonce code, a short arbitrary label such as \texttt{::DAX::} that the base model has never seen and assigns no meaning. The source prompts, answers, rendered target responses, response multiset, data split, optimizer, target modules, LoRA rank, update count, and evaluation support are identical across arms. Only the cue-to-mode relation changes.

Each arm and seed uses 1,928 training examples. The held-out factorial contains 468 events grouped into 117 source prompts. We train three seeds per arm on each base. At the first target-response token, $t^+$ is the predeclared target token and $t^-$ is the base model's highest-logit non-target token under one cue-free reference prompt shared across arms. The primary readout is common-base so the two independently trained descendants are compared in the same output coordinates; common-descendant and target-logit-only analyses are checks. The primary contrast, fixed before training, is $\C_{\mathrm{novel}}-\C_{\mathrm{familiar}}$.

\subsection{Uncertainty and aggregation}

Intervals are 95\% bootstrap confidence intervals. Released-checkpoint intervals resample prompt clusters within each model. Core-5 summaries first average within each family and then weight the five families equally. The controlled experiment uses a hierarchical bootstrap over training seeds and held-out source prompts; per-seed values are paired by source and mode. GSM8K intervals use paired prompt resampling. We report the unit and sample size with each detailed table in the appendix.

\section{Different Post-Training Runs Produce Different Dependencies}
\label{sec:results}

\subsection{Released descendants of one base span distinct profiles}

Start with the case that is hardest to anticipate. OpenMath2 is not a no-learning or no-late-effect control: on math prompts, swapping the base late stack for the OpenMath2 late stack moves the math-target margin by $+3.430$ logits after base upstream state and $+3.275$ after OpenMath2's own state, an interaction of $-0.154$ with an interval spanning zero. A large, genuinely learned, domain-specific capability sits in the late stack and works essentially as well on the earlier state of a checkpoint that never received the fine-tune. Effect magnitude and upstream dependence come apart \citep{toshniwal2024openmath}.

\begin{figure}[t]
  \centering
  \includegraphics[width=\linewidth]{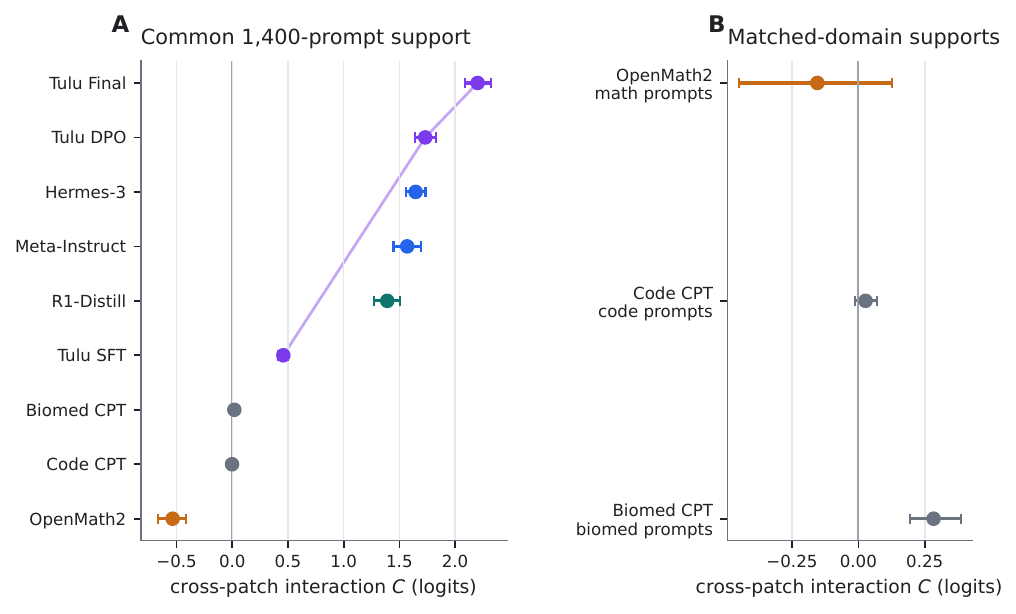}
  \caption{\textbf{Released and controlled descendants of the same Llama-3.1-8B base have different dependency profiles.} \textbf{A:} All points use the same 1,400-prompt support, layer-19 boundary, first-divergence event, and common-descendant readout. The line connects cumulative Tulu stages and is not a causal stage decomposition. \textbf{B:} Domain-matched supports. OpenMath2 retains a near-zero interaction despite a large math late-stack effect; biomedical CPT develops a smaller support-local interaction. Error bars are 95\% prompt-bootstrap intervals.}
  \label{fig:atlas}
\end{figure}

On the common 1,400-prompt support, Figure~\ref{fig:atlas}A places every same-base descendant on one axis. The interaction ranges from OpenMath2 at $-0.535$ logits to Tulu Final at $+2.202$. Meta Instruct, Hermes-3, R1-Distill, and the later Tulu stages are all strongly positive: their late changes lose much of their effect when fed base state. Tulu SFT is positive but smaller; connecting the three Tulu points shows accumulation along that released lineage without assigning the increments to SFT, DPO, or RLVR alone \citep{lambert2024tulu3,rafailov2023dpo}.

The CPT controls block a simpler explanation that any further training creates a positive interaction. Code and biomedical CPT improve held-out domain NLL by 4.66\% and 4.81\%, respectively, yet their interactions on the common prompt support are $-0.002$ and $+0.018$. Biomedical CPT reaches $+0.283$ on biomedical prompts, showing that dependence can be support-local rather than absent from the model. These controls are displayed next to, rather than folded into, the released atlas because their training and domain supports are deliberately narrower.

The atlas establishes variation, not cause: checkpoints differ in data, objectives, and training history. OpenMath2 and R1-Distill therefore do not isolate a ``math versus instruction'' mechanism. They show that neither a shared base nor a large late effect fixes $\C$. The controlled experiment isolates one training change.

\subsection{Changing only the cue relation shifts the dependency}

\begin{figure}[t]
  \centering
  \includegraphics[width=\linewidth]{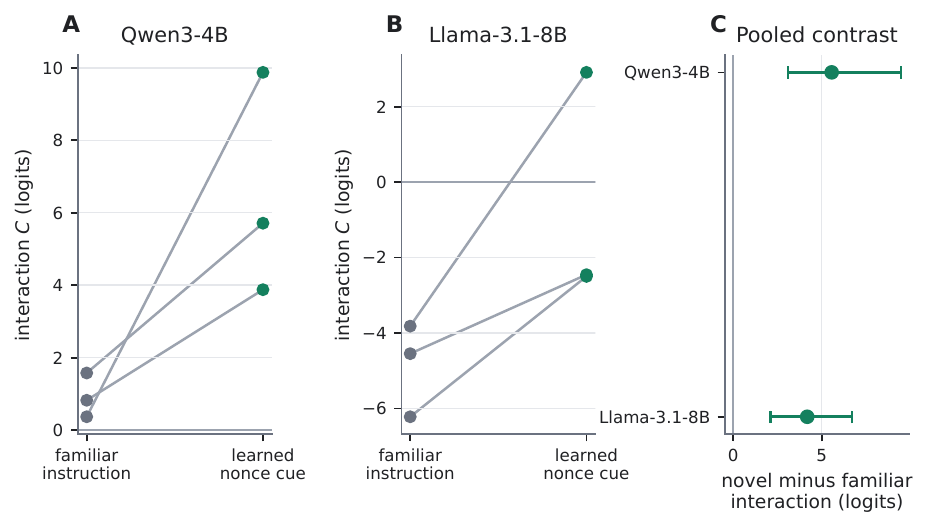}
  \caption{\textbf{Learning a new cue-to-response relation increases own-state dependence while target responses stay fixed.} \textbf{A--B:} Each line is one training seed, connecting the familiar natural-language instruction (gray) to the learned nonce cue (green). The target response set is identical across arms. \textbf{C:} Pooled novel-minus-familiar contrasts with hierarchical 95\% confidence intervals over seeds and held-out source prompts. The paired shift is positive for both bases and all six seeds.}
  \label{fig:controlled}
\end{figure}

Figure~\ref{fig:controlled} shows the controlled result. For Qwen3-4B, the novel-minus-familiar interaction is positive for every seed ($+3.053$, $+9.504$, and $+4.134$ logits) and is $+5.564$ when pooled across seeds (95\% CI $[+3.070,+9.467]$). For Llama-3.1-8B, all three paired shifts are also positive ($+2.094$, $+3.721$, and $+6.727$), with pooled effect $+4.181$ ($[+2.106,+6.698]$).

Both arms learned the requested behavior: held-out mode compliance is 99.8--100\% in every run, and exact-answer accuracy is similar within each base and seed. Effective LoRA update norms are tightly matched (9.19--9.71 on Qwen and 10.68--10.84 on Llama), and hash checks confirm identical source/mode supports and response multisets. Every factorial has 468 valid events and no invalid events. A comparator-free target-logit analysis remains positive when pooled: $+2.152$ on Qwen and $+4.129$ on Llama. Under descendant upstream state, the late replacement raises target-token top-1 more in the novel arm by 52.7 percentage points on Qwen and 41.0 points on Llama.

Holding targets and training setup fixed, replacing a familiar output instruction with a learned nonce-code mapping shifts the resulting late change toward greater dependence on its own upstream state. The absolute interaction remains architecture- and seed-dependent; two of three Llama novel-arm values are still negative. The result is a paired shift, not a claim that every novel cue produces a positive interaction or that every new behavior requires a new circuit. It directly shows that the relation learned during a fine-tune can change how earlier and later computation depend on each other even when the target responses are the same.

\section{Breadth and Consequence}

\subsection{The released instruction profile is broad}

\begin{figure}[t]
  \centering
  \includegraphics[width=\linewidth]{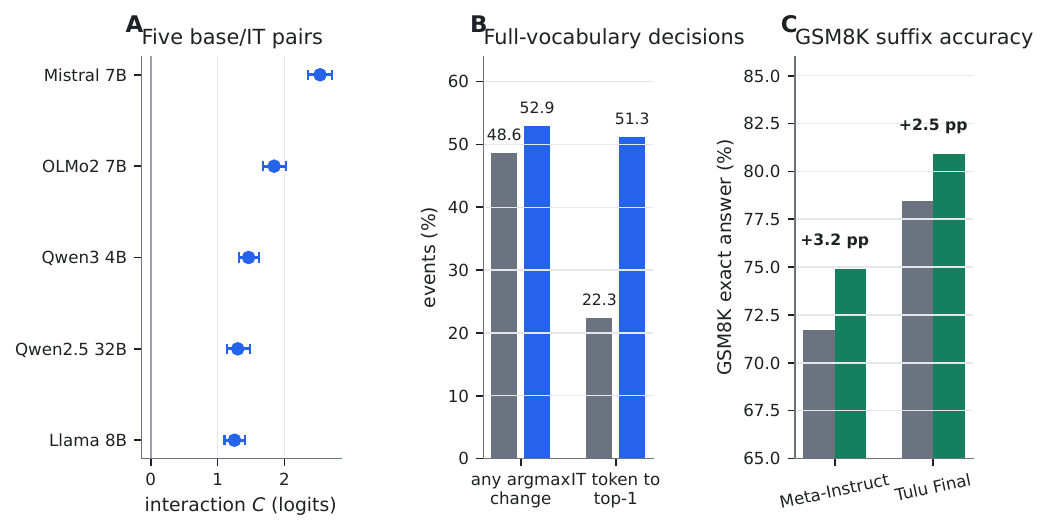}
  \caption{\textbf{The dependency is broad and reaches token decisions.} \textbf{A:} The interaction is positive in all five released base/instruction families; bars are 95\% prompt-bootstrap intervals. \textbf{B:} Replacing the base late stack with the instruction late stack changes the full-vocabulary argmax in about half of events under both base state (gray) and instruction state (blue), but promotes the instruction token to top-1 much more often under instruction state. \textbf{C:} On GSM8K, native instruction-token continuations (green) outperform continuations forced to begin with the base token (gray) for Meta Instruct and Tulu Final. The forced first token is excluded from exact-answer scoring.}
  \label{fig:breadth}
\end{figure}

The same interaction is positive in all five dense base/instruction pairs (Figure~\ref{fig:breadth}A). The family-balanced mean is $+1.680$ logits (95\% CI $[+1.603,+1.757]$). Replacing the base late stack with the instruction late stack adds $+0.759$ logits after base upstream state and $+2.439$ after instruction upstream state. Family interactions range from $+1.253$ for Llama to $+2.534$ for Mistral, with the 32B Qwen2.5 pair at $+1.302$. This result establishes breadth among the tested dense pairs; it is not a population estimate over all instruction-tuned or mixture-of-experts models.

\subsection{The local quantity reaches decisions}

The margin interaction is not merely a small logit perturbation. Under the common-descendant readout, replacing the late stack changes the full-vocabulary argmax in 48.6\% of events after base state and 52.9\% after instruction state (Figure~\ref{fig:breadth}B). It promotes the instruction token to top-1 in 22.3\% versus 51.3\% of events. The binary top-1 interaction is 30.9 percentage points. These are direct consequences of the four-cell intervention at the selected position.

We separately ask whether the chosen token can affect later correctness. On GSM8K \citep{cobbe2021gsm8k}, we locate the first content-bearing disagreement, force either the descendant-preferred or base-preferred token, continue with the native descendant, and score exact-answer correctness only on the suffix after the forced token. Across 1,120 valid Meta-Instruct events, the descendant-token branch scores 74.9\% versus 71.7\%, a paired gain of $+3.2$ points ($[+0.5,+5.8]$). Across 1,243 Tulu-Final events, the scores are 80.9\% versus 78.4\%, a gain of $+2.5$ points ($[+0.5,+4.5]$). The forced token is excluded from scoring.

The GSM8K experiment shows that a local divergent-token choice can have downstream objective consequences. It intervenes on the selected token rather than running the four hybrid cells as free-running models; Section~\ref{sec:discussion} states what that does and does not license.

\section{Focused Validity Tests}
\label{sec:validity}

The diagnostic relies on inherited coordinates, a chosen boundary, a selected token, and shared raw prefixes. We test the corresponding failure modes while leaving the full audit trail to Appendix~\ref{app:validity}.

\paragraph{Coordinates and boundary.}
The paired checkpoints share architecture and neuron indexing. A centered orthogonal Procrustes map fitted on half of each family's prompts leaves the held-out family-balanced interaction essentially unchanged: $+1.609$ logits before alignment and $+1.615$ after it, positive in all five families. The result is also positive at every tested boundary from 40\% depth through the final block, so it is not created by one global rotation or one specially chosen cut.

\paragraph{Token and history selection.}
Random later disagreements from the same rollouts retain 56\% of the first-divergence interaction, while scoring the eventual token pair before the models disagree retains only 3\%. Separate analyses after native instruction and base histories give $+1.51$ and $+1.49$ logits, respectively, and a fixed-continuation analysis remains positive through eight tokens ($+2.71$ at $N=8$). The effect is therefore concentrated at genuine disagreements but is not confined to one selected token or one raw shared history.

Appendix~\ref{app:validity} reports the boundary sweep, readout agreement, interpolation, signed-permutation, label-swap, and pre-late-commitment checks. Together, these controls make implementation failure, a global coordinate mismatch, one tuned boundary, and first-token selection inadequate explanations of the result. They do not make hybrid passes natural deployment trajectories.

\section{Related Work}

\paragraph{Where post-training changes a model.}
Work on post-training has identified localized or low-dimensional changes in knowledge, truthfulness, refusal, confidence, and instruction following \citep{du2025posttraining,arditi2024refusal,bigoulaeva2026patches}. Sparse model-diff methods likewise identify features that differ between base and chat checkpoints \citep{lindsey2024crosscoders,minder2025batchtopk}. These studies show where a learned behavior can be read out, controlled, or represented. Our question is different: whether the measured late change works after the base model's upstream state, and how that dependency changes across descendants and controlled training conditions.

\paragraph{Cross-model patching and fine-tuning mechanisms.}
Cross-model activation patching has been used to show that a math fine-tune strengthens an entity-tracking mechanism already present in its base \citep{prakash2024finetuning}. That work asks whether an existing task circuit is preserved and enhanced. We define a difference-in-differences quantity that compares the \emph{same} descendant late-stack replacement after base and descendant upstream states. The same-target experiment then changes the cue relation while holding target responses fixed. This turns upstream dependence from an observational description into a manipulated outcome.

\paragraph{Intervention validity.}
Activation patching can be misleading when states are off manifold, metrics are poorly chosen, or interventions open dormant pathways \citep{heimersheim2024patching,makelov2024illusion}. Our diagonal reconstruction, full-sequence recomputation, held-out alignment, boundary sweep, selection baselines, native histories, and readout swap target these risks. We retain the narrower interpretation that survives them: a local difference in how a late change uses upstream state.

\section{Discussion}
\label{sec:discussion}

The central result is not that transformer layers cooperate; they must. It is that post-training runs starting from the same base can leave measurably different dependencies between the learned late change and the state before it. The released atlas shows the range. The controlled experiment shows that the target responses alone do not determine the profile: changing only whether the output mode is requested by familiar language or a learned code shifts the interaction on two bases and every seed.

One useful interpretation is routing. A familiar instruction can already evoke a mode represented by the base model, allowing a fine-tuned late change to operate on a familiar upstream state. A new code must first be interpreted and associated with the requested mode, which can make the learned late change more dependent on the descendant's earlier computation. This interpretation is consistent with the controlled design, but $\C$ does not identify a circuit or prove that all novel skills behave this way. It is a testable account of one manipulated cue relation.

The released checkpoints suggest further hypotheses rather than settled recipe taxonomy. OpenMath2 may refine a math behavior for which the base upstream state already supplies useful candidates; general instruction post-training may require more context-dependent routing before late readout. The CPT controls show that generic domain learning is not sufficient to reproduce the instruction-support profile, while biomedical CPT shows that a dependency can emerge on its own domain. Because these recipes differ in many other respects, their positions should guide controlled follow-up experiments, not be read as causal labels.

This is why we recommend reporting $(\Delta_B,\C)$ alongside the native effect, as stated in Section~\ref{sec:intro}: the atlas shows that $\Delta_D$ on its own leaves the reader unable to tell OpenMath2's portable late effect from Tulu Final's state-dependent one, even though both are large.

\paragraph{Scope.}
First-divergence cross-patching measures local next-token compatibility, not average completion behavior or free-running model stitching. Native-history and constrained-continuation checks extend the pattern beyond one shared prefix and one scored token. The released atlas remains observational, and the controlled experiment covers two dense bases and a simple four-mode task. We therefore establish a local dependency diagnostic and a controlled cue-relation effect, not a recipe taxonomy or deployment-level transplantation result.

\section{Conclusion}

Cross-patching measures how strongly a learned late change depends on its descendant's earlier state. Released checkpoints differ on this dimension, and two fine-tunes trained on identical targets develop different profiles when only the cue-to-response relation changes. Post-training can therefore organize the same behavior through different dependencies across depth. A late-layer localization result does not by itself say which kind one is looking at, and the extra cell that settles it costs one forward pass.

\clearpage
\bibliographystyle{plainnat}
\bibliography{references}

\appendix

\paragraph{Appendix guide.}
Appendices~\ref{app:implementation}--\ref{app:models} document the intervention and evaluation supports; Appendices~\ref{app:atlas}--\ref{app:controlled} give the released and controlled results in full; and Appendices~\ref{app:validity}--\ref{app:reproducibility} report validity checks, decision-level analyses, and reproducibility details.

\section{Exact Cross-Patching Implementation}
\label{app:implementation}

\subsection{Event construction}

For released pairs, greedy generation starts from the same raw prompt. At each position we compare the base and descendant top-1 next tokens. The generated prefix remains identical until the first disagreement because the common token is appended whenever the models agree. The first position with different top-1 tokens defines the intervention event and the ordered token pair $(t^-,t^+)$. Events with tokenizer mismatch, invalid vocabulary entries, nonfinite logits, or no disagreement within the horizon are excluded before any effect is computed.

The raw-shared construction is an operational choice, not a claim about normal chat deployment. It guarantees identical token IDs and positions for residual replacement. Native-history controls use separately generated base and instruction histories and score local disagreements without requiring shared prefixes (Appendix~\ref{app:validity}).

\subsection{Hybrid forward passes and keys/values}

The cache used during greedy generation is discarded after locating the event. Every factorial cell is recomputed over the complete prompt-plus-prefix sequence with \texttt{use\_cache=False}. We run the upstream checkpoint through the selected boundary, replace the entire boundary hidden-state tensor, and run the selected late stack through the remaining blocks. The downstream model therefore computes its own attention keys and values for every prefix position from the patched hidden states. We do not feed an instruction late stack a cached base-model late-layer key/value tensor.

For $Y(U_B,L_B)$ and $Y(U_D,L_D)$, no-op hooks are applied through the same path as the hybrids. They match the corresponding native forward logits exactly in all retained records. Model-specific adapters resolve residual blocks, final normalization, output head, rotary state, and vocabulary masks.

\subsection{Readout convention}

The primary common-descendant readout applies the descendant final normalization, output head, and real-token mask to all four cell outputs. This prevents the output head from changing across cells. The common-base readout applies the same convention with the base readout. Core-5 interactions are $+1.680$ and $+1.704$ logits under the two conventions. Exp60 uses common-base as primary because the familiar and novel arms are separately trained descendants; common-descendant analyses agree.

\section{Models, Supports, and Boundaries}
\label{app:models}

\begin{table}[h]
\centering
\caption{Core-5 released base/instruction pairs and fixed late boundaries. Layer intervals are half-open.}
\label{tab:core-models}
\small
\begin{tabularx}{\linewidth}{lY Y c c}
\toprule
Family & Base checkpoint & Instruction checkpoint & Blocks & Late stack \\
\midrule
Llama-3.1 8B & Meta-Llama-3.1-8B & Meta-Llama-3.1-8B-Instruct & 32 & 19--31 \\
Qwen3 4B & Qwen3-4B-Base & Qwen3-4B & 36 & 22--35 \\
Mistral 7B & Mistral-7B-v0.3 & Mistral-7B-Instruct-v0.3 & 32 & 19--31 \\
OLMo-2 7B & OLMo-2-1124-7B & OLMo-2-1124-7B-Instruct & 32 & 19--31 \\
Qwen2.5 32B & Qwen2.5-32B & Qwen2.5-32B-Instruct & 64 & 38--63 \\
\bottomrule
\end{tabularx}
\end{table}

Pinned revision prefixes are, respectively: Llama \texttt{d04e592}/\texttt{0e9e39f}, Qwen3 \texttt{906bfd4}/\texttt{1cfa9a7}, Mistral \texttt{caa1feb}/\texttt{c170c70}, OLMo-2 \texttt{7df9a82}/\texttt{470b1fb}, and Qwen2.5 \texttt{1818d35}/\texttt{5ede1c9}. Full hashes and model licenses are included in the public artifact manifest. Model architecture sources are cited in the repository metadata \citep{dubey2024llama,qwen2025qwen3,jiang2023mistral,olmo2025olmo2,qwen2024qwen25}.

\paragraph{Core-5 support.}
The predeclared holdout contains 600 records: 300 assistant-conversation prompts, 150 explicit format-following prompts, and 150 safety prompts split evenly between harmful/refusal and safe-compliance cases. The valid-event counts in the paper synthesis are 600, 600, 597, 586, and 599 for Llama, Qwen3, Mistral, OLMo-2, and Qwen2.5. A later full-vocabulary audit recovered 598 rather than 599 Qwen2.5 records; Figure~\ref{fig:breadth}B uses the 598-record audit and Figure~\ref{fig:breadth}A uses the original pre-specified synthesis. This one-record difference does not affect any rounded conclusion.

\paragraph{Full 1,400-prompt support.}
The same-base atlas uses a fixed mixture of 300 factual questions, 200 reasoning questions, 250 format-following prompts, 300 assistant-conversation prompts, 150 safety prompts, 100 register/style prompts, and 100 easy baseline prompts. Factual and reasoning items draw from MMLU and GSM8K \citep{hendrycks2021mmlu,cobbe2021gsm8k}; format items draw from IFEval \citep{zhou2023ifeval}; the remaining categories use fixed committed manifests described in the artifact documentation. Hermes-3 and R1-Distill records were verified directly from run \texttt{exp59\_atlas\_full}: 1,400 source rows, \texttt{raw\_shared}, \texttt{first\_diff}, boundary 19, \texttt{full\_late}, and common-descendant readout. Their valid-event counts are 1,363 and 1,282.

\section{Released Same-Base Atlas and Domain Controls}
\label{app:atlas}

\begin{table}[h]
\centering
\caption{Llama-3.1-8B descendants on the common 1,400-prompt support. Tulu rows are cumulative checkpoints from one lineage.}
\label{tab:atlas}
\small
\begin{tabular}{lrrr}
\toprule
Descendant & $\C$ (logits) & 95\% CI & Valid events \\
\midrule
OpenMath2 & $-0.535$ & $[-0.670,-0.413]$ & 1,400 \\
Tulu SFT & $+0.458$ & $[+0.415,+0.499]$ & 1,316 \\
R1-Distill & $+1.391$ & $[+1.274,+1.507]$ & 1,282 \\
Meta Instruct & $+1.570$ & $[+1.447,+1.693]$ & 1,400 \\
Hermes-3 & $+1.646$ & $[+1.561,+1.735]$ & 1,363 \\
Tulu DPO & $+1.732$ & $[+1.641,+1.830]$ & 1,354 \\
Tulu Final & $+2.202$ & $[+2.091,+2.322]$ & 1,362 \\
\bottomrule
\end{tabular}
\end{table}

\begin{table}[h]
\centering
\caption{Domain controls distinguish a small interaction from a missing learned late effect.}
\label{tab:domain}
\small
\begin{tabularx}{\linewidth}{lY rrr}
\toprule
Model/support & Learned-domain check & $\Delta_B$ & $\Delta_D$ & $\C$ \\
\midrule
OpenMath2 / math & released math SFT; math prompts & $+3.430$ & $+3.275$ & $-0.154$ \\
Code CPT / common prompts & held-out code NLL improves 4.66\% & --- & --- & $-0.002$ \\
Biomed CPT / common prompts & held-out biomed NLL improves 4.81\% & --- & --- & $+0.018$ \\
Code CPT / code prompts & matched domain support & --- & --- & $+0.027$ \\
Biomed CPT / biomed prompts & matched domain support & --- & --- & $+0.283$ \\
\bottomrule
\end{tabularx}
\end{table}

The CPT adapters are controlled non-instruction continuation fine-tunes from the same pinned Llama base. Both pass merge checks and have 0\% degenerate generations in the health audit. Their purpose is not to match the scale of a released instruction pipeline, but to show that measurable domain learning and same-base weight drift do not automatically yield the common-support interaction.

\section{Controlled Same-Target Experiment}
\label{app:controlled}

\subsection{Data construction}

Each source contains a short problem and canonical answer. The four response modes are:
\begin{itemize}
  \item \textbf{plain}: the canonical answer;
  \item \textbf{JSON}: one object with a \texttt{final\_answer} string;
  \item \textbf{XML}: the answer wrapped in a fixed XML field;
  \item \textbf{reverse}: the canonical answer string reversed.
\end{itemize}
The familiar arm gives an ordinary output instruction. The novel arm gives one stable code per mode (for example, \texttt{DAX} for one mode). The code assignment is fixed before training. Every source appears in every mode, and construction tests assert byte-identical target response multisets and identical source/mode supports across arms. Training loss is applied only to response tokens.

\begin{table}[h]
\centering
\caption{Controlled experiment summary. Counts are per arm and training seed.}
\label{tab:exp60-design}
\small
\begin{tabularx}{\linewidth}{lYY}
\toprule
Property & Familiar arm & Novel arm \\
\midrule
Cue & natural-language mode instruction & fixed nonce code \\
Target responses & \multicolumn{2}{c}{identical plain/JSON/XML/reverse response multiset} \\
Training rows & 1,928 & 1,928 \\
Held-out factorial & 468 events / 117 sources & 468 events / 117 sources \\
Seeds & 60, 61, 62 & 60, 61, 62 \\
Compliance & 99.8--100\% & 99.8--100\% \\
\bottomrule
\end{tabularx}
\end{table}

\subsection{Per-seed results}

\begin{table}[h]
\centering
\caption{Controlled interaction by base and training seed. The pre-specified estimand is the paired novel-minus-familiar column.}
\label{tab:exp60-seeds}
\small
\begin{tabular}{lrrrr}
\toprule
Base & Seed & Familiar $\C$ & Novel $\C$ & Novel $-$ familiar \\
\midrule
Qwen3-4B & 60 & $+0.825$ & $+3.877$ & $+3.053$ \\
Qwen3-4B & 61 & $+0.370$ & $+9.874$ & $+9.504$ \\
Qwen3-4B & 62 & $+1.577$ & $+5.712$ & $+4.134$ \\
Llama-3.1-8B & 60 & $-4.547$ & $-2.453$ & $+2.094$ \\
Llama-3.1-8B & 61 & $-6.220$ & $-2.499$ & $+3.721$ \\
Llama-3.1-8B & 62 & $-3.817$ & $+2.910$ & $+6.727$ \\
\bottomrule
\end{tabular}
\end{table}

The common-descendant readout reproduces the pooled estimates. Using the target token logit without a comparator gives $+2.152$ ($[+0.060,+5.045]$) on Qwen and $+4.129$ ($[+2.111,+5.627]$) on Llama. Under descendant upstream state, the novel-minus-familiar increase in target-token top-1 effect is $+52.7$ points ($[+44.3,+61.1]$) and $+41.0$ ($[+25.0,+71.1]$), respectively.

\paragraph{Analysis amendment.}
The initial arm-specific comparator selected the base model's top alternative under each arm's prompt. Because the familiar and nonce prompts differ, those alternatives matched in only 2 of 468 paired events. Before pooling, we replaced this with the base model's top non-target token under one cue-free reference prompt shared across arms. Targets, trained adapters, boundaries, and all hybrid hidden states were unchanged. Llama seed 60 was rescored under this shared comparator and is explicitly selected by the analyzer; all other runs were produced after the correction. The target-logit-only result above is comparator-free.

\section{Breadth and Validity Tables}
\label{app:validity}

\begin{table}[h]
\centering
\caption{Core-5 common-descendant results. Each interval resamples prompts within family.}
\label{tab:core5}
\small
\begin{tabular}{lrrr}
\toprule
Family & Valid events & $\C$ & 95\% CI \\
\midrule
Llama-3.1 8B & 600 & $+1.253$ & $[+1.104,+1.417]$ \\
Qwen2.5 32B & 599 & $+1.302$ & $[+1.137,+1.489]$ \\
Qwen3 4B & 600 & $+1.464$ & $[+1.315,+1.619]$ \\
OLMo-2 7B & 586 & $+1.847$ & $[+1.674,+2.026]$ \\
Mistral 7B & 597 & $+2.534$ & $[+2.349,+2.718]$ \\
\midrule
Family-balanced & 2,982 & $+1.680$ & $[+1.603,+1.757]$ \\
\bottomrule
\end{tabular}
\end{table}

\begin{table}[h]
\centering
\caption{Cross-family boundary sweep on a fixed event set. All five family interactions are positive at every boundary.}
\label{tab:boundaries}
\small
\begin{tabular}{lrrr}
\toprule
Late boundary & Family-balanced $\C$ & 95\% CI & Family range \\
\midrule
40\% depth & $+1.667$ & $[+1.576,+1.765]$ & $[+0.393,+3.153]$ \\
50\% depth & $+1.795$ & $[+1.706,+1.885]$ & $[+0.788,+3.070]$ \\
60\% depth & $+1.692$ & $[+1.608,+1.781]$ & $[+0.538,+3.161]$ \\
70\% depth & $+1.418$ & $[+1.358,+1.479]$ & $[+0.886,+2.129]$ \\
80\% depth & $+1.146$ & $[+1.097,+1.197]$ & $[+0.624,+1.712]$ \\
90\% depth & $+0.843$ & $[+0.800,+0.887]$ & $[+0.376,+1.473]$ \\
Final block & $+0.340$ & $[+0.310,+0.370]$ & $[+0.154,+0.538]$ \\
\bottomrule
\end{tabular}
\end{table}

\begin{table}[h]
\centering
\caption{Central validity checks and the failure mode each addresses.}
\label{tab:controls}
\small
\begin{tabularx}{\linewidth}{lY Y}
\toprule
Check & Result & Interpretation \\
\midrule
Held-out Procrustes & inherited $+1.609$; aligned $+1.615$ & not explained by one global orthogonal mismatch \\
Random later disagreement & 56\% of first-divergence interaction & high-signal selection, not a unique token \\
Future pair before divergence & 3\% of first-divergence interaction & ordered token labels do not create the effect early \\
Native instruction history & $+1.51$ $[+1.42,+1.61]$ & survives without raw shared history \\
Native base history & $+1.49$ $[+1.33,+1.65]$ & same-sign mirror \\
Common-base readout & $+1.704$ vs. $+1.680$ primary & not an output-head convention \\
Signed permutations & 0.31$\times$ endpoint magnitude & magnitude alone is insufficient \\
Label swap & $p=5\times10^{-5}$ & orientation is not reproduced by relabeling \\
\bottomrule
\end{tabularx}
\end{table}

\paragraph{Constrained continuation.}
The continuation support pools four Llama instruction descendants on the full 1,400-prompt manifest. For each valid first-divergence event, we force the descendant token, construct short native descendant and base candidate tails, and teacher-force both tails through the four cells. The sequence score is the descendant-candidate log probability minus the base-candidate log probability. At $N=0$, $\C=+1.50$ ($[+1.45,+1.56]$, 5,432 events). At $N=8$, $\C=+2.71$ ($[+2.59,+2.84]$, 3,422 events). On the same $N=8$ survivor subset, the $N=0$ interaction is $+1.17$, so horizon growth is not solely event dropout. A same-forced descendant-tail control is $+2.46$; a shuffled-tail control is $+0.87$. This establishes fixed-continuation persistence, not natural rollout behavior.

\section{Decision and GSM8K Details}
\label{app:behavior}

\begin{table}[h]
\centering
\caption{Family-balanced full-vocabulary decision rates under the common-descendant readout.}
\label{tab:decisions}
\small
\begin{tabular}{lrr}
\toprule
Late-stack replacement outcome & Base upstream & Instruction upstream \\
\midrule
Any argmax change & 48.6\% $[46.9,50.3]$ & 52.9\% $[51.2,54.7]$ \\
Instruction token promoted to top-1 & 22.3\% $[20.9,23.8]$ & 51.3\% $[49.5,53.0]$ \\
Mean $\log_2$ rank improvement & $+0.735$ $[+0.690,+0.781]$ & $+0.964$ $[+0.919,+1.014]$ \\
\bottomrule
\end{tabular}
\end{table}

For GSM8K, prompts without a content-bearing base/descendant disagreement are excluded from the primary valid-event analysis; an all-prompt analysis assigns them zero treatment difference and remains positive. The branch continuation is greedy under the native descendant after forcing one token. The exact-answer parser sees only the suffix after the forced token. For Meta Instruct, native-descendant-token accuracy is 74.91\% and forced-base-token accuracy is 71.70\% ($n=1{,}120$, exact paired $p=0.0181$). For Tulu Final the corresponding values are 80.93\% and 78.44\% ($n=1{,}243$, $p=0.0178$). Rank-matched alternative branches and content-token subsets are included in the released summary.

\section{Reproducibility, Compute, and Licenses}
\label{app:reproducibility}

Code, compact result summaries, claim checkers, and figure scripts are available at
\begin{center}
\href{https://github.com/yifan1207/first-divergence-crosspatching}{\texttt{github.com/yifan1207/first-divergence-crosspatching}}.
\end{center}
The repository separates three reproduction levels: (1) CPU claim checking from committed JSON/CSV summaries; (2) figure and table regeneration; and (3) optional multi-GPU raw intervention reruns. The arXiv source bundle contains this manuscript, the official NeurIPS style file used for typesetting, vector figures, and the bibliography. A companion supporting-material archive contains the compact evidence roots for the retained paper claims and a SHA-256 manifest.

The broader research program used approximately 2,000 GPU-hours, including exploratory and paper-excluded analyses. The retained experiments use A100/H100-class GPUs for released-model cross-patching and LoRA training; CPU synthesis regenerates all reported tables and figures from compact outputs. The artifact manifest records run-level hardware and approximate duration so this paper does not imply that a full raw rerun is CPU-cheap.

Model and dataset licenses are recorded per artifact. In brief, Llama checkpoints use the Llama 3.1 Community License; Qwen checkpoints use Apache-2.0; Mistral v0.3 uses Apache-2.0; OLMo-2 uses Apache-2.0; Tulu artifacts use their released model/data terms; OpenMathInstruct-2 is released under its dataset terms; GSM8K and MMLU are MIT-licensed; and IFEval follows its repository license. Users must separately accept gated model terms where applicable.

\end{document}